\documentclass[letterpaper, 10 pt, journal, twoside]{IEEEtran} 
\IEEEoverridecommandlockouts                             
\usepackage{xspace}
\usepackage{times}
\usepackage{amsfonts}
\usepackage{verbatim}
\usepackage{graphicx}
\usepackage{subfigure}
\usepackage{multirow}
\usepackage{graphics}
\usepackage{latexsym,amsfonts,amssymb}
\usepackage[noline,noend,ruled,linesnumbered]{algorithm2e}
\usepackage{amsmath}
\usepackage{fancyhdr}
\usepackage{wrapfig}
\usepackage{subfigure}
\usepackage{setspace}
\usepackage{scrextend}
\usepackage[bookmarks=true]{hyperref}

\title{A Dataset for Improved RGBD-based Object
  Detection and Pose Estimation for Warehouse Pick-and-Place }

\author{Colin Rennie$^{1}$, Rahul Shome$^{1}$, Kostas E. Bekris$^{1}$, and Alberto F. De Souza$^{2}$%
\thanks{Manuscript received: September, 1, 2015; Revised December, 22, 2016; Accepted January, 30, 2016.}%Use only for final RAL version
\thanks{This paper was recommended for publication by Editor Dr. Jana Kosecka upon evaluation of the Associate Editor and Reviewers' comments. 
  The authors would like to thank the sponsors of Rutgers
  University' participation to the Amazon Picking Challenge: Yaskawa
  for providing a Motoman SDA10F robot, UniGripper for providing a
  custom-made vacuum gripper, Robotiq for providing a three-fingered
  hand, Amazon for providing the shelving unit, the objects
  associated with the challenge and a modest travel award.}%Use only for final RAL version
\thanks{$^{1}$Computer Science, Rutgers University, Piscataway, New
  Jersey, USA. Email:\url{kostas.bekris@cs.rutgers.edu}}%
\thanks{$^{2}$Computer Science, Federal University of Espirito Santo,
  Brazil. Email: \url{alberto@lcad.inf.ufes.br}}%
\thanks{Digital Object Identifier (DOI): see top of this page.}
}
\begin{document}
\maketitle
% \thispagestyle{empty}
% \pagestyle{empty}

%%%%%%%%%%%%%%%%%%%%%%%%%%%%%%%%%%%%%%%%%%%%%%%%%%%%%%%%%%%%%%%%%%%%%%%%%%%%%%%%

% KOSTAS

\begin{abstract}
An important logistics application of robotics involves manipulators
that pick-and-place objects placed in warehouse shelves. A critical
aspect of this task corresponds to detecting the pose of a known
object in the shelf using visual data. Solving this problem can be
assisted by the use of an RGBD sensor, which also provides depth
information beyond visual data. Nevertheless, it remains a challenging
problem since multiple issues need to be addressed, such as low
illumination inside shelves, clutter, texture-less and reflective
objects as well as the limitations of depth sensors. This paper
provides a new rich dataset for advancing the state-of-the-art in
RGBD-based 3D object pose estimation, which is focused on the
challenges that arise when solving warehouse pick-and-place tasks.
The publicly available dataset includes thousands of images and
corresponding ground truth data for the objects used during the first
Amazon Picking Challenge at different poses and clutter conditions.
Each image is accompanied with ground truth information to assist in
the evaluation of algorithms for object detection. To show the utility
of the dataset, a recent algorithm for RGBD-based pose estimation is
evaluated in this paper. Given the measured performance of the
algorithm on the dataset, this paper shows how it is possible to
devise modifications and improvements to increase the accuracy of pose
estimation algorithms. This process can be easily applied to a variety
of different methodologies for object pose detection and improve
performance in the domain of warehouse pick-and-place.

\end{abstract}

% Keywords appear just beneath the abstract. Use only for final RAL version.  

%%%%%%%%%%%%%%%%%%%%%%%%%%%%%%%%%%%%%%%%%%%%%%%%%%%%%%%%%%%%%%%%%%%%%%%%%%%%%%%%

% Introduction:
	% Pick & place in warehouse environment using popular RGB-D sensors
	% 	Define importance and properties of environment, e.g. semi-structed
	% 	Talk briefly about competition properties - in particular work_order
	% Dataset allowing researchers to further tailor pose estimation approaches to such an environment, while evaluating enviornmental factors such as clutter and position

% Background:
	% Other RGBD datasets: properties, gt labels, environments, controls (e.g. structured, semi-structured)
	% Perhaps monocular/stereo datasets as well...
	% Do we talk about other pose estimation approaches here??

\section{Introduction}
\label{sec:intro}
% Drop letter for first word of the Introduction 
% Here we have the typical use of a "T" for an initial drop letter
% and "HIS" in caps to complete the first word.
\IEEEPARstart{T}{here} is significant interest in warehouse automation, which
frequently involves pick-and-place tasks for products located in
shelving units. 
This interest is exemplified by competitions such as
the first Amazon Picking Challenge (APC) \cite{Correll:2016aa}, which
brought together multiple academic and industrial teams from around
the world, as well as similar competitions, like the Robocup@Home
Challenge \cite{robocup_home}.  One way to approach the APC involved
perception, motion planning and grasping of 25 different objects,
which were placed in a semi-structured way inside the bins of an
Amazon-Kiva Pod. Solving such problems reliably can significantly
alter the logistics of distributing products. Frequently, manipulation
research on pick-and-place tasks has focused on flat surfaces, such as
tabletops. These are relatively simpler problems, which do not involve
many of the issues that often arise in warehouse automation, where the
presence of tight spaces, such as shelves, plays a critical role.

\begin{figure}[t]
\centering
\includegraphics[width=0.48\textwidth]{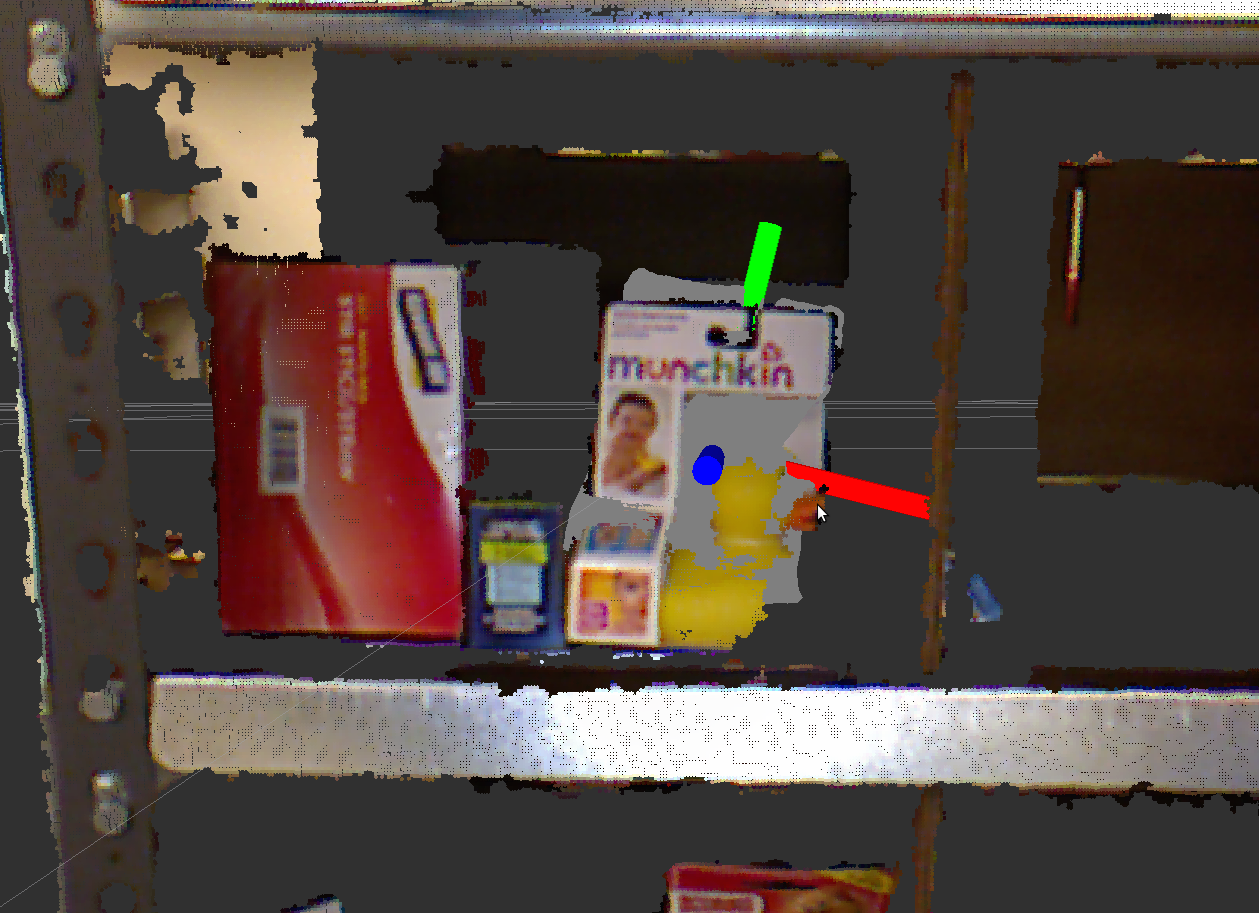} 
% \vspace{-0.2in}
\vspace{-0.1in}
\caption{An example frame from the Rutgers dataset, where a pose estimate
generated by the test algorithm is superimposed.  \vspace{-0.4in}
}
\label{fig:munchkin_detection}
\end{figure}

Accurate pose estimation is crucial for successfully picking an object
inside a shelf. In flexible warehouses, this pose will not be a priori
known but must be detected from sensors, especially visual ones. The
increasing availability of RGBD sensors, which can simultaneously
sense color and depth, brings the hope that such problems can be
eventually solved reliably. But warehouse shelves have narrow, dark and
obscuring bins that complicate object detection. Clutter can further
challenge detection through the presence of multiple objects. A
variety of object types need to be dealt with, some of which may be
texture-less and not easily identifiable from color, and others
reflective and virtually undetectable by a depth sensor.  Furthermore,
some popular depth sensors exhibit limits in terms of the smallest and
largest sensing radius that make it harder for a manipulator to
utilize them. Thus, RGBD-based object detection and pose estimation is
an active research area and a critical capability for warehouse
automation.

This paper provides tools that help in improving the performance of
object detection solutions for such challenges. In particular, it
describes a new rich dataset and software for utilizing it. The
motivation is to better equip the research community in evaluating and
improving robotic perception solutions for warehouse picking. The
dataset contains over 10,000 depth and RGB registered images, complete
with hand-annotated 6DOF poses for 24 of the APC objects (for details,
see Section 3). Also provided are 3D mesh models of the APC objects,
which may be used for training of recognition algorithms. The code for
utilizing and integrating the dataset with different algorithms is
also publicly available.

The dataset includes images of warehouse objects in a shelf
environment. The objects are placed in different poses in various bins
of warehouse shelves, so as to allow a variety of experimental
conditions (Figure \ref{fig:shelf_positions}). Multiple camera
perspectives and frames account for rich information, as well as
spatial and temporal variation in data. The effect of clutter is
evaluated by controlling the presence of additional objects in a
scene.

The dataset is compared against the one available as part of the
LINEMOD framework for object detection \cite{linemod1}, to highlight
the need for additional varying conditions, such as clutter, camera
perspective and noise, which affect pose detection. This is the chief
contribution of the dataset, the utility of which is further evaluated
by using the open-source implementation of the LINEMOD
framework \cite{linemod1}) easily accessible via
OpenCV \cite{Bradski:2000aa}.  This paper does not argue that this
algorithm is the best solution for pose estimation in shelves. The
method is used as an example of a modern, accessible algorithm for
object detection, which at least performs effectively in tabletop
setups.

The dataset reveals that the considered algorithm faces significant
difficulties when used in a warehouse scenario. This allows to
appreciate the features of warehouse picking, which complicate pose
estimation. With the aid of the dataset, it was also possible to
identify algorithmic and engineering adaptations to increase
performance in warehouse pick-and-place.

% The paper
% provides the success rate for detecting objects in the APC shelves
% within predefined thresholds and under different clutter conditions in
% an incremental manner, i.e., as the various modifications are
% introduced to the baseline LINEMOD framework. Most of these changes
% are agnostic to the internal operations of the pose detection
% algorithm and can be applied across different methods.

Overall, the proposed dataset emphasizes the need for the development
of pose detection algorithms that can operate robustly for a wide
variety of objects and conditions, especially in narrow, dark and
cluttered spaces. Such algorithms need to optimally utilize all
available sources of sensing data and prior information.

% Add in object models
% Show a picture of setup environment, pictures of objects used

\section{Related Work}
\label{sec:background}
% Describe datasets generally (properties, etc)

\begin{figure*}
	\centering
	\begin{minipage}{.80\textwidth}
		\centering
		\includegraphics[width=.99\textwidth]{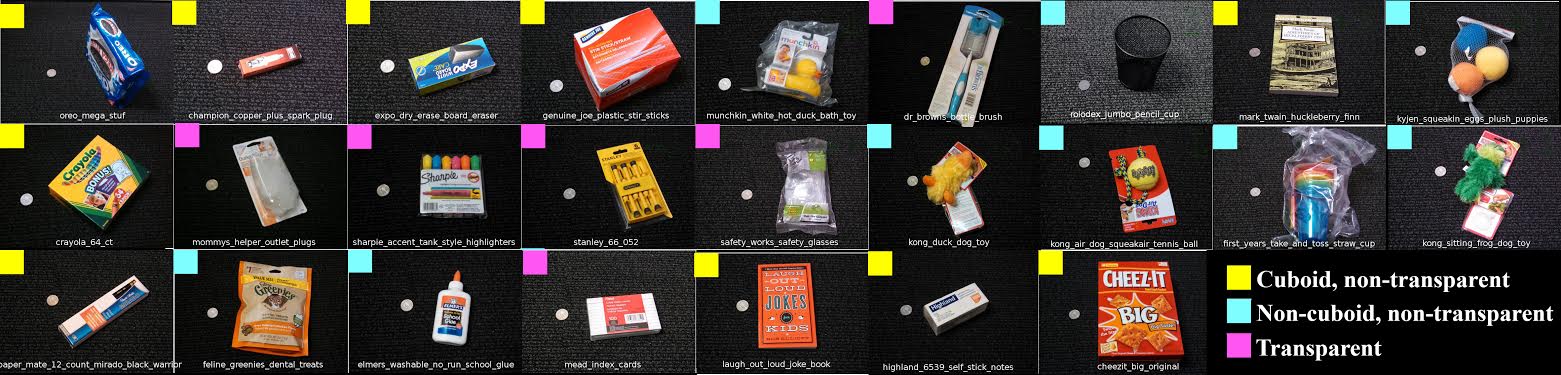}
		\label{fig:objectlist}
	\end{minipage}%
	\begin{minipage}{.19\textwidth}
		\centering
        \vspace{0.15in}
		\includegraphics[width=.99\textwidth, height=0.14\textheight]{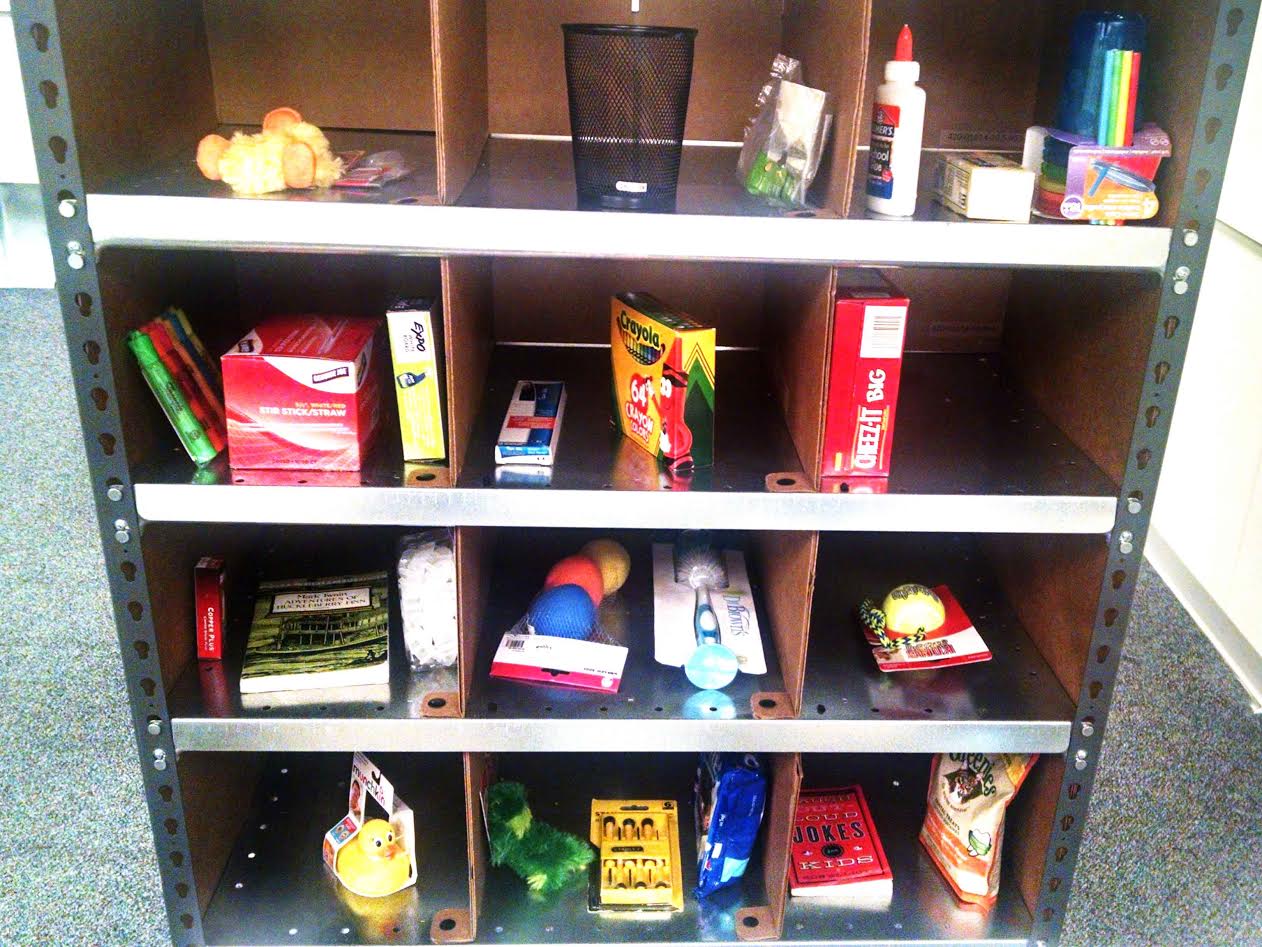}
		\label{fig:shelf}
	\end{minipage}
    \vspace{-0.15in}
	\caption{(Left) Items used in the Amazon Picking Challenge
          2015 and featured in the dataset. Three groups of objects
          are identified based on their effects on pose estimation
          from RGBD data: a) cuboid and non-transparent, b)
          non-cuboid and non-transparent, c) transparent. (Right) An
          arrangement of the shelf with the APC objects.}
	\label{objects}
    \vspace{-0.1in}
\end{figure*}

Datasets for the task of object recognition have rapidly grown in
recent years both in terms of number as well as size and scale. The
applications of such datasets include industrial warehouse
applications like the APC \cite{Correll:2016aa}, and domestic
applications like in the RoboCup@home
\cite{stuckler2012demonstrating}. Some standard RGB benchmarks for the
task include CIFAR-10/100 \cite{cifar}, ImageNet \cite{imagenet}, and
PASCAL VOC \cite{pascalvoc}. Some use bounding boxes as ground truth
and others use image segmentation with inliers/outliers for accuracy
metrics. While useful for 2D image object recognition, RGB datasets
are not ideal in manipulation applications, which rely not only on
segmenting the object of interest but also on accurate pose
estimation.

Up until the last decade, the problem of 3D recognition was often
addressed using a stereo camera. More recently, RGBD cameras'
availability and widespread use have increased interest in solutions
to common 2.5D\footnote{2.5D refers here to the projection of a 2D
  image to 3D space, which results in a sparse 3D image.} problems,
such as face, object, and gesture recognition. Such technology has
allowed researchers to begin to build ``modern-scale'' datasets, which
help evaluating performance and identifying challenges. Several such
datasets are described below\footnote{A more complete list of
  available RGBD datasets can be found at:
  \url{http://www0.cs.ucl.ac.uk/staff/M.Firman/RGBDdatasets/} }.\\

\noindent\textit{Segmented Scenes Datasets}

\textbf{B3DO \cite{b3do}: } A project by UC Berkeley, the dataset
contains $>$3,000 2.5D crowd-sourced images. The images primarily
focus on indoor scenes, where ground truth bounding boxes have been
annotated for more than 50 object categories. The dataset has also
been augmented to include $(x,y,z)$ Cartesian coordinates for many
object centroids.

\textbf{NYU Depth Dataset v2 \cite{nyu}: } The NYU dataset also
focuses on indoor scenes, but ground truth labels are presented as
full image segmentation. The dataset includes around 500k 2.5D
images, with approximately 1,500 fully labeled ground truth images. \\

\noindent\textit{Manipulation Datasets}

\textbf{YCB Objects \& Models \cite{ycb}: } A collaboration between
several robotics labs, the YCB dataset provides object models in a
variety of formats for common household objects. The focus of this
project is to create common metrics for the growing interest in
robotic manipulation research by providing reliable benchmarks for
several common manipulation tasks.\\

\noindent\textit{Object Datasets}

\textbf{Table-Top Object Dataset \cite{tabletop}: } A collaboration
between Willow Garage and the Univ. of Michigan, this dataset consists
of $\sim$1,000 2.5D images with ground truth labels for 480
frames. The objects presented belong to 3 different classes, each
class consisting of approximately 10 different instances. Objects are
shown on table tops in clutter of between 2-6 items per image. The
images were collected using a structured light stereo camera.

\textbf{Solutions in Perception Dataset \cite{solutionsinperception}:
} This dataset by Willow Garage contains 35 objects in $\sim$1,000 3D
training images and 120 test images. In training, objects were
presented in clutter with 6DOF ground truth for each item.  The scenes
were captured using RGBD cameras with objects on a turn-table to
capture and reconstruct the scenes from multiple viewpoints. All
images were captured using a consistent azimuth angle between the
camera and the turntable.

% In the test dataset, only a single item is presented with ground truth. 

\textbf{UW Dataset \cite{uw}: } This large dataset consists of over 50
object categories and 300 distinct instances. It features objects from
multiple viewpoints, and is presented with ground truth pose for one
axis [0,2$\pi$].

\textbf{LINEMOD Dataset \cite{linemod_accv}: } As part of the body of
work detailing the LINEMOD framework, the authors released a dataset
of 18 object models and over 15,000 6D ground truth annotated RGBD
images. Objects in these images are shown in clutter from a variety of
viewpoints. Because of the size, setting, and focus on 6D pose
estimation, this dataset is the most closely related to the current
paper. \\

The dataset proposed here presents more than 10,000 ground truth
annotated RGBD images of 24 objects of different types. As opposed to
prior datasets \cite{uw,nyu,b3do}, this new dataset is specifically
aimed at perception for robotic grasping and hence features full 6DOF
ground truth poses for all 2.5D images.  While some existing datasets
\cite{tabletop,linemod_accv} provide ground truth poses for objects in
cluttered space, the new one additionally controls for clutter by
presenting poses of the objects both with and without clutter. Other
controls employed in data collection correspond to multiple viewpoints
and collection of additional frames for control of slight noise in
sensors. Additionally, scenes are not reconstructed as in alternatives
\cite{solutionsinperception}, but the dataset includes the
transformation matrices between the camera location, stationary
robotic base, and object location. This allows users of the dataset to
reconstruct the scene to suit their own methods. Lastly, this new
dataset is specifically designed for warehouse perception task and is
focused on the placement of objects in narrow spaces, such as shelf
bins. To the best of the authors' knowledge, this is the first attempt
to generate a real-world dataset for this important application.

%A significant feature of this dataset is the control with which it was
%collected and the metrics these controls will allow users to define.

% Describe basic use case of LINEMOD in using the dataset
\section{Rutgers APC RGBD Dataset}
\label{sec:dataset}
This paper presents a large 2.5D dataset consisting of 10k+ images and
corresponding ground truth 6DOF poses for all these images, which is
made available to the research community \footnote{It can be accessed
  online at the following url:
  \url{http://www.pracsyslab.org/rutgers_apc_rgbd_dataset}}. The focus
is on supporting warehouse pick-and-place tasks. The accompanying
software allows the easy evaluation of object detection and pose
estimation algorithms in this context.

\subsection{Objects and 3D-mesh Models}

The selected objects correspond to those that were used during the
first Amazon Picking Challenge (APC) \cite{Correll:2016aa}, which took
place in Seattle during May 2015.
% The same is true for the shelving unit,
% which is the one provided by Amazon for the purposes of the
% competition. 
% Figure \ref{objects} (right) depicts an example object
% configuration inside of the Amazon-Kiva Pod shelf. The set of APC objects that were part of the competition exhibit representative characteristics in the problem domain, including different size, shape, texture and transparency.
% exhibit good variety in terms of various characteristics, such as
% size, shape, texture, transparency and are good candidates for
% objects that need to be transported by robotic units in warehouses.

The provided dataset comes together with 3D-mesh object models for
each of the APC competition objects. For most objects, the CAD 3D
scale models of the objects were textured using the open-source
MeshLab software. For simple geometric shapes, such as cuboids, this
simple combination of CAD modeling and texturing is sufficient and can
yield results of similar quality to more involved
techniques \cite{narayan2015range}. For objects with non-uniform
geometries, models were produced using 3D photogrammetric
reconstruction from multiple views of a monocular camera.

%\begin{figure*}[t]
%	\centering
%	\includegraphics[width=0.9\textwidth]{images/meta_items.jpg}
%	\label{fig:items}
%	\caption{}
%\end{figure*}

\subsection{Dataset Design}

The intention with the dataset was to provide to the community a large
scale representation of the problem of 6DOF pose estimation using
current 2.5D RGBD technology in a cluttered warehouse shelf
environment.  The set of 25 APC objects that were part of the
competition exhibit a variety of features in the problem domain,
including different size, shape, texture and transparency.

% The objects are placed in a variety of different poses, from multiple viewpoints. Every viewpoint also has multiple frames to account for effects of noise. 

% This involved representing the challenge in
% such a way that would allow researchers to determine the effects of
% several parameters to success ratio and accuracy, such as the effects
% of clutter and object types.

% items of clutter. In this way, the dataset allows one to parse out the
% degree to which environmental clutter affects accuracy. The dataset
% presents these varying clutter scenarios for each item inside of each
% of the 12 bins within the shelving unit.

% Additionally, while rotating each object and accompanying clutter
% items throughout the bins of the Kiva Pod, the pose remains consistent
% within each bin. The pose of an object changes, however, each time the
% object is placed in a different bin. This ensures that the set of
% chosen poses represent good coverage of likely positions for the
% objects of interest.

\begin{figure}[t]
	\centering \includegraphics[width=0.48\textwidth]{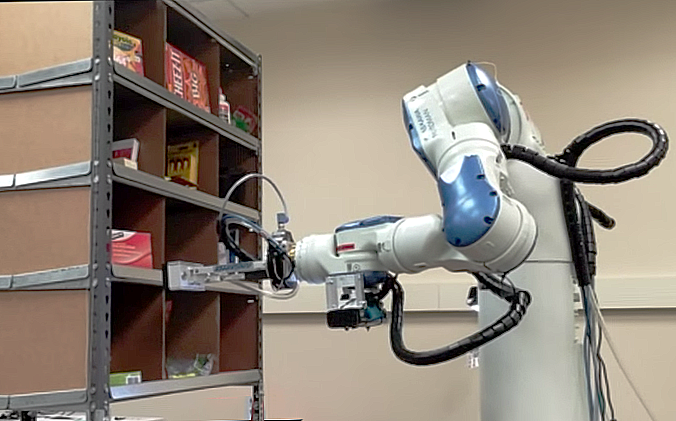} 
	\vspace{-0.1in} 
	\caption{The
	data collection setup for the warehouse pick-and-place
	dataset: A Motoman SDA10F robot and an Amazon-Kiva Pod stocked
	with objects. At this configuration, the Kinect sensor mounted
	on the arm is used to detect an object at the bottom row of
	shelf bins. \vspace{-0.1in}} \label{fig:datacollectionsetup}
\end{figure}

% \subsection{Extent of the Dataset}
\subsection{Extent of the Dataset}

Data collection was performed using a Microsoft Kinect v1 2.5D RGBD
camera mounted to the end joint of a Motoman Dual-arm SDA10F robot
(Figure \ref{fig:datacollectionsetup}). Changing the intensity of the
structured light in the Kinect driver allowed operation at closer
distances.

% Two LED lighting
% strips were added to the camera so as to control the illumination of
% the environment across images. The position of the camera was
% calibrated prior to data collection to ensure accurate transformations
% between the base of the robot, the camera, and the detected and
% annotated ground truth poses. 

% Ask Chuples how far away our mapping positions are from the center of the bin 
% And which mapping positions correspond to which positions

To provide better coverage of the scene and the ability to perform
pose estimation from multiple vantage points, data from 3 separate
positions (referred to, here, as ``mapping'' positions) were
collected: i) One directly in front of the center of a bin at a
distance of 48cm, ii) a second roughly 10cm to the left of the first
position, and iii) a third with the same distance to the right of the
first position.  Four 2.5D images were collected at each mapping
position to account for noise.

To measure the effects of clutter, for each object-pose combination,
images were collected: (1) with only the object of interest occupying
the bin, (2) with a single additional item of clutter within the bin,
and (3) with two additional items of clutter. In all, the dataset can
be broken down into the following parameters:

\begin{itemize}

\item 24 Objects of interest\footnote{The
          ``mead\_index\_cards'' item from the APC list is not
          included as this simplified the experimental process for
          collecting the data and it was the item that exhibited the
          most redundant qualities.}
\item 12 Bin locations per object
\item 3 Clutter states per bin
\item 3 Mapping positions per
	clutter state
\item 4 Frames per mapping position
\end{itemize}

Considering all these parameters, the dataset is composed of a total
of 10,368 2.5D images. For each image, there is a YAML file available
containing the transformation matrices (rotation, translation)
between: (1) the base of the robot and the camera, (2) the camera and
the ground truth pose of the object, and (3) the base of the robot and
ground truth pose of the object.

%%ground truth generation

The process for generating the ground truth data involved iterating
over all the frames of the Rutgers APC dataset in a semi-manual
manner. A human annotator translated and rotated the 3D model of the
object in the corresponding RGBD point cloud scene using RViz. Every
annotation superimposes the model to the corresponding portion of the
point cloud.\footnote{Additional details regarding naming conventions
for the dataset and instructions for download and use can be found at
the project's
website: \url{http://www.pracsyslab.org/rutgers_apc_rgbd_dataset}.}

% The pose of the object model that best matched the RGBD
% point cloud is saved as the ground truth, in addition to the
% transformations and frames needed to regenerate the scene. For the
% cases of incomplete or noisy point cloud data, the color images were
% used as a complementary reference point.

% \subsection{Naming Conventions}

% Files within the dataset follow a strict naming convention in order to
% be easily parsed based on researchers' needs. The naming convention is
% as follows:\\

% \vspace{-.15in}
% \textit{[obj\_name]\-[f\_type]\-[bin]\-[clutter]\-[map\_pos]\-[frame]}\\
% \vspace{-.15in}
% \begin{addmargin}[1em]{1em}
% \textbf{obj\_name: } the name of the object, using APC naming
% conventions\\
% \textbf{f\_type:} file type; any one of \{image, depth, pose\}\\
% \textbf{bin:} \{A-L\} corresponding to the top-to-bottom right-to-left locations of the bins in the shelf\\
% \textbf{clutter:} \{1-3\} indicates the number of items in the bin of interest (including target)\\
% \textbf{map\_pos:} camera mapping position; any of \{1-3\}\\
% \textbf{frame:} any of \{1-4\}\\
% \end{addmargin}
% \vspace{-.15in}
% Files are distributed in .png format aside from the pose YAML file.

% Describe environment, background of use of LINEMOD, examples of reasons enhancements needed

\section{Opportunities for Pose Estimation Improvements Through the Dataset}
\label{sec:evaluation}

% A key concept of the APC, which mirrors situations that can arise in
% real warehouses, is that of the ``work order'', which provides: a) a
% list of items to be gathered, b) their bin locations in the shelf
% unit, c) along with the locations of all other items contained in the
% warehouse shelf.  Given the work order, a target object needs to be
% detected and its pose estimated robust enough so as to allow the
% picking of the corresponding item by a manipulator without disturbing
% neighboring items and placing it in an order bin. This is a
% semi-structured task, where the system has partial information about
% which items can be expected to lie within certain areas but no further
% information about how the objects will appear within that area. Thus,
% the key question that needs to be addressed from a perception point of
% view is that of pose estimation.

This paper employs a setup similar to that of the Amazon Picking
Challenge (APC) to evaluate the proposed dataset, which is a helpful
testing ground for robotic perception algorithms in a relatively
controlled but realistic warehouse environment. The available software
infrastructure for using the dataset allows the incorporation of
different algorithms for this problem, given the rich literature on
the subject \cite{Ma:2015aa, Ma:2014aa, Choi:2012aa, Choi:2012ab,
Lai:2011aa, Aldoma:2011aa, Pham:2011aa, Muja:2011aa, Kim:2011aa,
Steder:2010aa, Drost:2010aa, Triebel:2010aa, Collet:2009aa}. The
current paper utilizes one such approach that is easily accessible to
the robotics community and corresponds to the LINEMOD
algorithm \cite{linemod1}, for which an implementation based on the
OpenCV library
\cite{Bradski:2000aa} is available.  
% The objective is to demonstrate
% the difficulty of the perception problem in the context of warehouse
% pick-and-place as well as highlight critical aspects that need to be
% considered for effectively addressing it. The use of the dataset in
% conjunction with the LINEMOD algorithm also led to the development of
% a series of pre- or post-processing steps that improve accuracy, which
% can be potentially applied to different algorithms as well.

% \begin{figure*}[t]
% 	\centering
% 	\includegraphics[width=1\textwidth]{images/results}
% 	\caption{Positive detection and pose estimation results
%           expressed as \% pose estimations under our chosen threshold
%           of 5cm/15deg evaluated over the entire Rutgers APC dataset
%           for four variants of the original LINEMOD software. For
%           illustration purposes of the use case for this dataset, we
%           only use 12 of the 24 total objects. Black line shows
%           averages across 12 object categories. }
% 	\label{fig:results_plot}
% \end{figure*}

LINEMOD is an object detection and pose estimation
pipeline \cite{linemod1}, which received as input a 3D mesh object
model. From the model, various viewpoints and features from multiple
modalities (RGB gradients, surface normals) are sampled. The features
are filtered to a robust set and stored as a template for the object
and the given viewpoint. This process is repeated until sufficient
coverage of the object is reached from different viewpoints. The
detection process implements a template matching algorithm followed by
several post-processing steps to refine the pose estimate. The
approach was designed specifically for texture-less objects, which are
notoriously challenging for pose estimation methods based on color and
texture. LINEMOD uses surface normals in the template matching
algorithm and limits RGB gradient features to the object's silhouette.

Starting with the \textit{baseline} open source implementation of
LINEMOD, the paper shows the incremental performance improvements
achieved over the basic implementation algorithm through the use of
the Rutgers APC dataset.  Most of the improvements are
algorithm-agnostic and can be useful in general to warehouse detection
and pose estimation tasks.

% \subsection{Baseline}

% This paper performs an evaluation of the baseline LINEMOD approach on
% the dataset using the corrected
% OpenCV\cite{Bradski:2000aa}/ORK\cite{ork} implementation of the
% algorithm, as described in the previous section.  When applied to
% warehouse pick-and-place, the performance of the algorithm has several
% limitations. 
% % The focus on texture-less objects means that the method
% % relies heavily on the shape and contours of the object to compute a
% % similarity score between stored object templates and the contours of
% % the image \cite{Hinterstoisser:2013aa}. When this information is
% % sparse, as often the case when objects are placed in shelves, the
% % algorithm's employment of an ICP \cite{Fitzgibbon:2003aa} variant for
% % pose refinement can cause pose estimates to vary between frames. 
% % In
% % the OpenCV/ORK implementation, a single inlier-to-outlier threshold
% % must be defined to determine positive matches. An ideal choice for
% % this value may vary based on which object is being identified in the
% % image or other factors, such as lighting conditions. 
% The following
% performance improvements to the framework are tailored to the
% warehouse picking challenge.

\subsection{Masking}

% A picking problem on a factory floor will involve a more or less fair
% knowledge about the general location of the object in the work
% order. For instance, i
In the context of the APC, the bin of the shelf
from which the object is detected and grasped is specified. In order
to take advantage of such information, precise calibration of the
shelf's location with respect to the robot is performed prior to
detection. Using ROS' TF functionality \cite{ros} it is possible to
compute the boundary of the current bin of interest:
$([x_{min},x_{max}], [y_{min},y_{max}], [z_{min},z_{max}])$. Then, all
points $p_i = (p_i^x,p_i^y,p_i^z)$ are masked if
$$\prod_{j\in(x,y,z)}(j_{max}>p_{i}^{j}>j_{min}) == 0 \vspace{-0.05in}$$

\subsection{Post-processing}
% Mentioned in the baseline evaluation is the problem of a single
% threshold not adapting well to different conditions. 

Dynamically selecting a viable detection threshold value allows the
algorithm to always allow a fixed number of detections through and
increases the positive detection rate in the context of warehouse
pick-and-place since it is known that the object is present in the
scene. Pose detection is made more accurate by dividing the RGB image
into four quadrants and doing a hue-saturation histogram comparison
for individual quadrants. The method is similar to an existing
one \cite{hshist} with the addition of quadrant processing, which aids
in predicting the correct orientation of the object.

% An additional step of post-processing leverages the color information
% of the RGBD dataset. However, hue-saturation histogram comparisons
% evaluated on the entire image fail when the object faces have similar
% color distribution. This happens often in warehouse items which are
% covered with commercial packaging with similar color schemes. 

\subsection{Temporal smoothing}
A single query for object detection operates over a single frame of
RGBD data from a single perspective. Capturing multiple frames from
the same perspective helps in mitigating effects of noisy sensor data
or inconsistent pose estimates. In the implementation of the temporal
smoothing enhancement, 12 frames of RGBD images are aggregated, with
the most frequently reported pose estimate reported on the final
frame. Effectively, for all $P_i \in P_{pose\_estimations}$:
$$\underset{P_i} {\mathrm{argmax}} (Q(P_i) + \sum_{\forall ~P_j \in
neigh(P_i)} Q(P_j)/dist(P_i,P_j) )$$
\noindent where $Q(P_i)$ is a quality measure of pose estimation
$P_i$, $dist(P_i,P_j)$ is a distance function between poses, and
$neigh(P_i)$ returns all other pose estimates within a small
neighborhood of $P_i$.

% Getting images from different perspectives gives the detector a better
% chance to get an accurate estimate and smoothing the final result over
% multiple detections generally improves the robustness of the overall
% pose estimation for objects. One caveat is that, 

For objects where the likelihood of getting a good detection is low,
temporal smoothing might bias the final detection towards bad pose
estimations. Nevertheless, in this environment the positive effects of
smoothing pose estimates outweigh the negatives on average.

\section{Features and Comparison}
\label{sec:results}

One of the defining characteristics of this dataset is the amount of
control put into isolating certain environmental factors in its
collection. To exemplify the importance of these controls, an analysis
over the example LINEMOD algorithm is performed over both the proposed
dataset and the original LINEMOD dataset.

\begin{figure}[t]
	\includegraphics[width=0.49\textwidth]{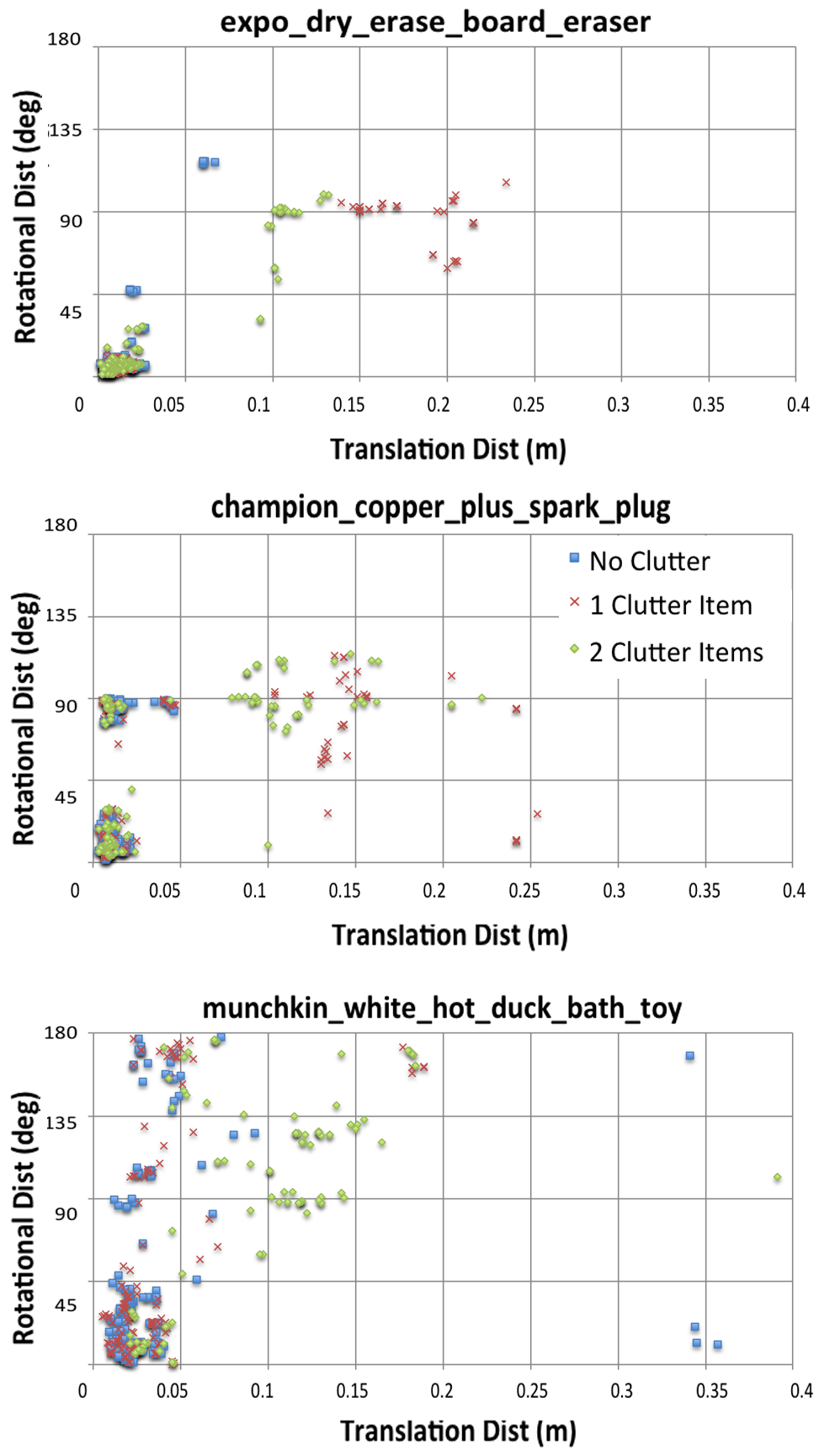}
	\vspace{-0.3in}
	\caption{Scatter plots of raw pose estimation accuracy results for three example objects from the APC dataset. X-axis is translational error (L2 dist) in meters, Y-axis is rotational error in degrees. }
	\label{fig:scatter_big}
    \vspace{-0.15in}
\end{figure}

\begin{figure}[t]
	\includegraphics[width=0.49\textwidth]{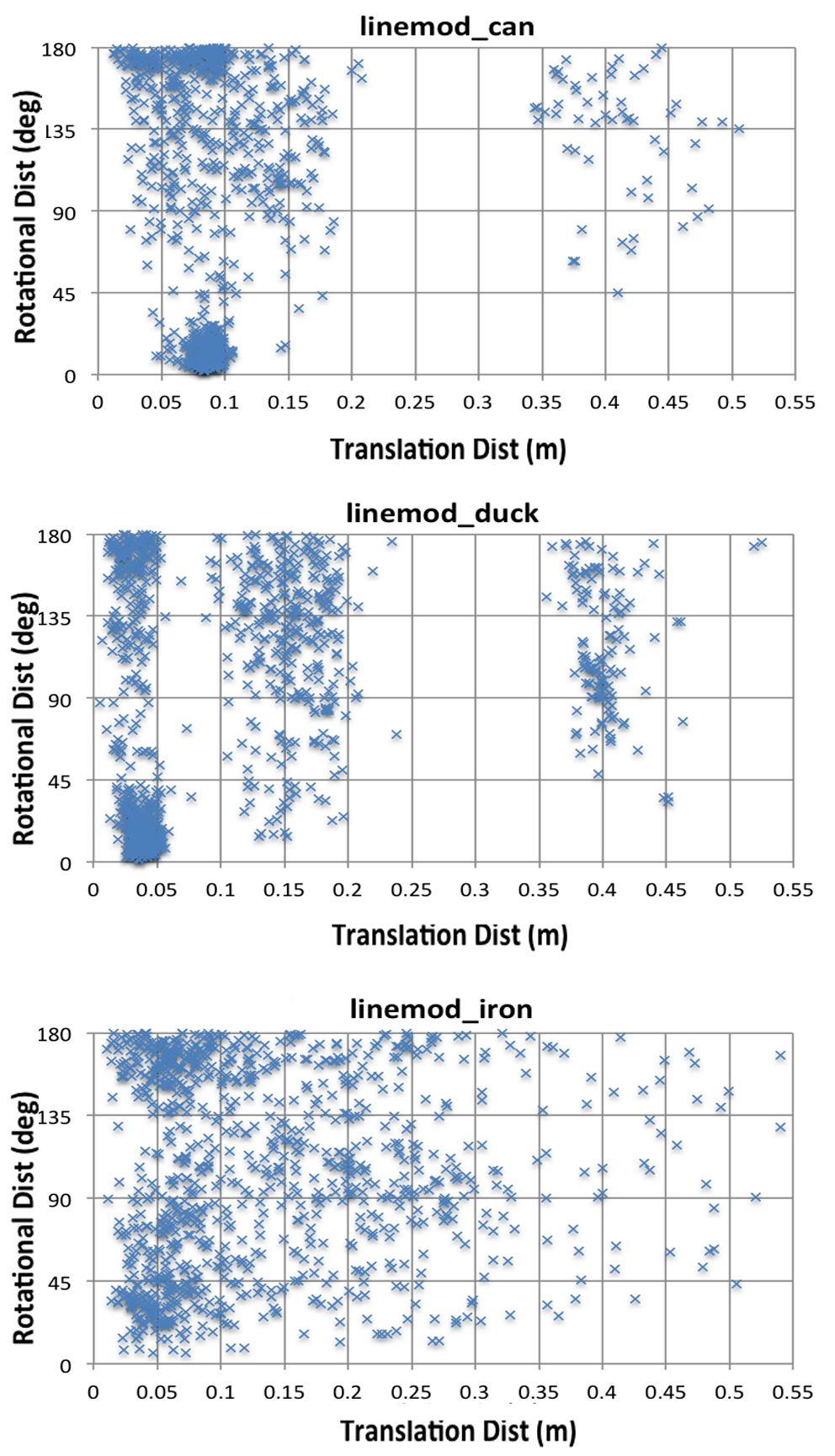}
    \vspace{-0.3in}
	\caption{Scatter plots of raw pose estimation accuracy results for three example objects from the LINEMOD dataset. Axes measure the same dimensions as the plots to the left. }
	\label{fig:linemod_results}
    \vspace{-0.14in}
\end{figure}

\subsection{Effects of Clutter}

A situation known to cause difficulty for pose estimation algorithms,
including state-of-the-art solutions, corresponds to the presence of
significant clutter present in the scene containing the target
object. For example, in a scene containing only the target object,
simple segmentation techniques may provide reliable results. Adding,
however, even a small number of other objects with vaguely similar
colors or other visual features can easily cause simple approaches to
fail. As such, it is a high priority goal for current solutions to 6D
pose estimation problems to be as robust as possible to the presence
of clutter.

To elaborate on the description of the control for clutter, for every
distinct target object pose across each of the 12 bins, the dataset
provides frames (i) with the object alone in the bin, (ii) accompanied
by a single clutter item, and (iii) accompanied by two clutter
items. By doing so, users of the dataset can directly compare the
accuracy of different algorithms under these three different
conditions.

Though the LINEMOD dataset shares similarities with the one proposed
here (e.g., in terms of providing 6D ground truth, a variety of scenes
and poses, and environments containing lots of clutter), it does not
provide insights regarding the effects of clutter. In the comparison
provided in this paper on Figures \ref{fig:scatter_big} and
\ref{fig:linemod_results}, these insights are exemplified.  In the
graph from Figure \ref{fig:scatter_big}, which corresponds to the
proposed dataset, a majority of the inaccurate pose estimates arise
from scenes containing more clutter, but the overall effect is
relatively small. In the plot from Figure \ref{fig:linemod_results},
corresponding to the LINEMOD dataset, inaccurate pose estimates occur across all
variations in clutter and are dominated by the translation error in
cluttered scenes. This is a likely indication that the detection
algorithm is confusing other objects for the object of interest and
thus indicates a weaker detection strength for this target
object. Specifically regarding the middle graph of Figure
\ref{fig:linemod_results}, there is a cluster of inaccurate pose
estimates at the 90-degree rotational error rate. Due to the cuboid
shape of the target object, it is likely that the algorithm often
estimates an incorrect orientation of the object. Made possible by the
controls placed in the data collection of the proposed dataset, these
types of observations are valuable when making improvements to pose
estimation algorithms or when comparing different approaches to suit a
specific task. Additionally, given the inconsistent accuracy observed
when evaluating the LINEMOD dataset, these insights would be extremely
useful in this case.

\subsection{Coverage and Variety of Poses}

\begin{figure}[t]
	\centering
        \includegraphics[width=0.45\textwidth]{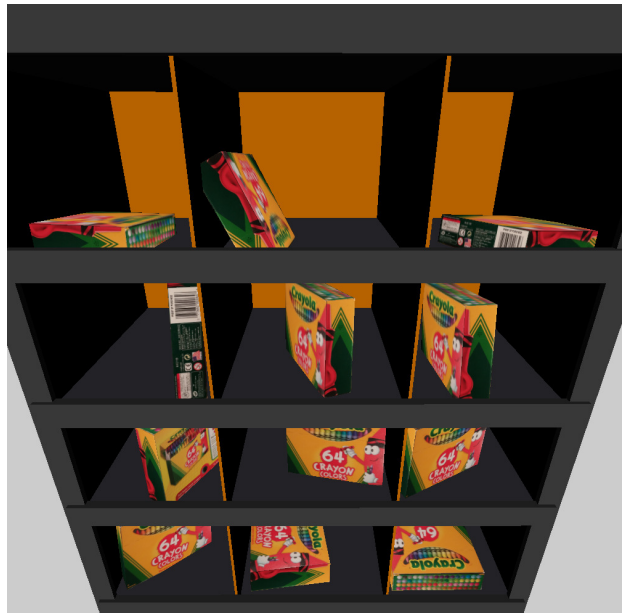}
        \vspace{-.1in}
	\caption{Simulated scene showing the variety of ground truth
          poses for one example object from the proposed dataset, as
          it is rotated through the 12 bins of the shelf. }
    \vspace{-0.1in}
	\label{fig:shelf_positions}
\end{figure}

Another control prioritized when assembling the proposed dataset was
to ensure a variety of ground truth poses for each object. At the same
time, there was also the objective to choose poses such that they were
representative of probable placements of objects in the APC
competition. As such, all target placements are located several
centimeters away from the front edge of the bin. The utility of the
coverage characteristic is in allowing users to test their solutions
when each of the major faces of the object is the primary viewable
face. This allows to immediately identify troublesome surfaces and,
given these insights, to design more robust solutions. Figure
\ref{fig:shelf_positions} illustrates this control for one example
object.

\subsection{Viewpoint Variety}

Among the additional features of the proposed dataset is the
accumulation of samples from several vantage points for each target
object pose and clutter combination. Specifically, samples were
collected for each configuration from three different viewpoints in
front of the shelf: (i) left of, (ii) centered in front of, and (iii)
right of the bin. Since the left and right positions may incur some
level of occlusion of the target object by parts of the shelving unit,
one of the applications of this feature is in the determination of the
effects of these partial occlusions. Additionally, these samples can
be used for pose hypothesis aggregation and smoothing, or for 3D
reconstruction approaches.

\subsection{Noisy Sensing}

While Kinect v1 is an inexpensive and widely available sensor, a major
detriment in its use is the noise inherent in RGBD samples produced
using this equipment. To counteract this, for each configuration and
camera position, the dataset provides four samples taken over a period
of several seconds with all objects and hardware stationary. Similar
to the above, this feature will allow users to easily determine which
situations and target objects are robust to this noise and which are
not.

\subsection{Extensions}

In addition to the above controls, an inherent feature of the dataset
is that it can be used not only for single-object pose estimation, but
also for multi-object. Because all transforms from object to robotic
base are stored in the ground truth pose files, users may easily
extend this dataset to the multi-object case simply by reading in all
ground truth poses of neighboring bins within the same ``run'', or
configuration, of data collection. Since within a single ``run'', no
item's placement is changed, this is straightforward to do. And
because the dataset is organized by these runs, the implementation is
rather easy. This feature makes the dataset a good candidate for
testing 3D reconstruction techniques.

% What we learned
% Shortcomings of warehouse/linemod, need for improvement
% Talk about use of our statistics as a starting point, hope to improve
\section{Discussion}
\label{sec:discussion}
% Excluded objects include ones where the RGBD sensor data is heavily
% compromised due to transparent, shiny or mesh-like surfaces. Depth
% based detection schemes suffer from this limitation of RGBD
% sensors. The unique challenges in object detection and pose
% estimation that show up are highlighted in the evaluation results.

% Rutgers APC dataset provides an unique opportunity to examine the
% pose estimation problem in the context of warehouse
% pick-and-place. The size of the dataset coupled with the presence of
% annotated ground truth poses for all RGBD frames opens up avenues
% for implementing and evaluating pose estimation algorithms to work
% on the dataset. As a further contribution of this paper, a LINEMOD
% based object detection and pose estimation solution is implemented
% and improved upon. Although there exists algorithms that might
% provide better results than the current implementation, the current
% work identifies several improvements to RGBD object detection and
% pose estimation that are agnostic to the algorithm being used. These
% include masking, histogram based

This work contributes a large hand-annotated RGBD dataset with 6DOF
ground truth poses. The dataset is specifically designed to support
advancing solutions for the problem of pose estimation in tight
environments that appear in warehouse picking problems.  The extent
and structure of the dataset provides flexibility to researchers and
allows them to use the data to apply and evaluate pose estimation
methods using a variety of different techniques. The dataset is not
only large relative to alternatives but is also designed to allow
evaluation of several additional factors that can affect pose
estimation accuracy. The accompanying software allows for improvements
that are agnostic to the pose detection algorithm.

The evaluation of an easily available pose estimation algorithm to the
robotics community over the proposed dataset emphasizes the
difficulties that RGBD-based solutions face when dealing with
transparent and reflective surfaces \cite{Lysenkov:2012aa}. Cuboid
objects also pose some difficulties for algorithms that are based
primarily on RBG-D data but it was possible to deal with these issues
through the improvements described in this work, which were tailored
to fit the context of a warehouse environment and provide robustness.

%The contributions of this paper are provided to the community in the
%hope that they can establish a unified platform for evaluating new
%techniques that both take advantage and address the difficulties of
%RGBD data in the context of warehouse picking.

The current dataset does not focus on the case ofpartially occluded
objects, where a pose estimation process may be used to evaluate the
pose of both the occluding and the occluded objects so as to assist
rearrangement manipulation algorithms \cite{Krontiris:2015aa,
Krontiris:2016aa}.  Such problems can be potentially benefited by the
utilization of cloud computation in order to improve performance and
deal with the inferent uncertainty in the pose estimation and
manipulation processes \cite{Bekris:2015aa}.

There is also an influx of new results in the area of machine learning
that can potentially be applied for the problem of pose estimation for
warehouse picking and it would be interesting to see the quality of
such solutions given the available dataset. Similarly, more classical
methods developed for monocular cameras that primarily depend upon
color and texture may exhibit complementary behavior to the one
displayed by the considered RGBD approach. Fusing such methods can
also be another way of achieving solutions that operate robustly over
a wide variety of object classes and environmental conditions. The
dataset can be useful in the evaluation of solutions in the context of
related applications, such as directly detecting a
handle \cite{Ten-Pas:2014aa} or a grasp \cite{Ten-Pas:2015aa} from
point cloud data.

\bibliographystyle{IEEEtran}
\bibliography{perception}

\end{document}